\documentclass[11pt]{article}

\usepackage[margin=1in]{geometry}
\usepackage[T1]{fontenc}
\usepackage[utf8]{inputenc}
\usepackage{lmodern}
\usepackage{microtype}
\usepackage[english]{babel}

\usepackage{amsmath}
\usepackage{graphicx}
\usepackage{subcaption}
\usepackage{float}
\usepackage{booktabs}
\usepackage{longtable}
\usepackage{hyperref}

\usepackage[backend=biber,style=authoryear,maxcitenames=2,sortlocale=en_US,language=english]{biblatex}
\title{compar:IA: The French Government's LLM arena to collect French-language human prompts and preference data}
\author{Lucie Termignon \\ Simonas Zilinskas \\ Hadrien P{\'e}lissier \\ Aur{\'e}lien Barrot \\ Nicolas Chesnais \\ Elie Gavoty \\[0.5em] \texttt{contact@comparia.beta.gouv.fr}}
\date{}

\newcommand{\snapshotdate}{2026-02-07}

\begin{document}
\sloppy
\hbadness=10000
\maketitle

\begin{abstract}
Large Language Models (LLMs) often show reduced performance, cultural alignment, and safety robustness in non-English languages, partly because English dominates both pre-training data and human preference alignment datasets. Training methods like Reinforcement Learning from Human Feedback (RLHF) and Direct Preference Optimization (DPO) require human preference data, which remains scarce and largely non-public for many languages beyond English.

To address this gap, we introduce \textbf{compar:IA}, an open-source digital public service developed inside the French government and designed to collect large-scale human preference data from a predominantly French-speaking general audience. The platform uses a blind pairwise comparison interface to capture unconstrained, real-world prompts and user judgments across a diverse set of language models, while maintaining low participation friction and privacy-preserving automated filtering.

As of \snapshotdate, compar:IA has collected over 600,000 free-form prompts and 250,000 preference votes, with approximately 89\% of the data in French (platform analytics; snapshot \snapshotdate). We release three complementary datasets---conversations, votes, and reactions---under open licenses, and present initial analyses, including a French-language model leaderboard and user interaction patterns.

Beyond the French context, compar:IA is evolving toward an international digital public good, offering reusable infrastructure for multilingual model training, evaluation, and the study of human--AI interaction.
\end{abstract}

\section{Introduction}

Large language models (LLMs) are overwhelmingly trained on English-language data. Publicly available training disclosures for major open models show that French typically represents a very small share of pre-training data, with similarly low proportions during post-training stages such as instruction tuning and preference alignment. For example, the Llama 2 technical report indicates that French accounts for only 0.16\% of the training corpus \parencite{touvron2023llama2}.

This imbalance causes degraded fluency, mismatched register, culturally inappropriate responses, and weaker safety guarantees in non-English languages \parencite{conneau2020xlmr,hershcovich2022crosscultural,bigoulaeva2021crosslingual,wang2024alllanguages}. Preference data captures human judgments crucial for training techniques like RLHF and DPO \parencite{christiano2017preferences,ouyang2022instructgpt,rafailov2023dpo}, but large-scale preference datasets remain rare outside English, especially as open resources.

While proprietary systems likely collect large volumes of multilingual interaction data, these datasets are generally inaccessible to smaller or emerging industrial, academic and public-sector actors, reducing model diversity and competition.

Blind pairwise comparison has emerged as an effective method for collecting scalable human preference data while reducing brand and expectation biases \parencite{chiang2024arena}. LLM arenas, most notably the LMSYS Chatbot Arena, now known as Arena\footnote{\url{https://arena.ai/}}, have demonstrated that crowdsourced preferences can meaningfully contribute to model training and evaluation. However, participation and data generation remain heavily concentrated in English, which constrains evaluation and training efforts for other languages. Public release policies also differ across arenas, and only a subset of collected interactions is typically released as open data.

To address these gaps, the French government launched \textbf{compar:IA} (\url{https://comparia.beta.gouv.fr/}), a public LLM arena designed to collect human preference data from a predominantly French-speaking audience. compar:IA adapts blind pairwise comparison for a general public by removing barriers to entry and providing clear explanations, while remaining compatible with established preference-based evaluation practices.

Since its public launch in October 2024 and as of \snapshotdate, compar:IA has collected over 600,000 free-form prompts and more than 250,000 preference votes (platform analytics; snapshot \snapshotdate). All published data are released continuously under Etalab 2.0 licenses on Hugging Face and data.gouv.fr, following a privacy filtering pipeline that excludes conversations containing detected personal data. Based on publicly released resources, compar:IA appears to be among the largest openly available collections of French-language human prompt and preference data for conversational AI. Additionally, since November 2025, the compar:IA interface has been adapted to other languages, paving the way for the creation of datasets in other less resourced languages.

This paper documents compar:IA as both an awareness-raising platform and a data collection infrastructure. We describe the platform's design and user experience, the structure and publication of the resulting datasets, and early indicators of adoption and impact. We also discuss limitations related to user representativeness and task coverage, and outline future directions, including multilingual expansion. More broadly, this work contributes empirical evidence to ongoing discussions on multilingual data, evaluation, alignment, and the role of public institutions in building open AI infrastructure.

\section{The compar:IA platform}

\subsection{User interaction flow}

The interaction flow consists of a small number of clearly separated steps:

\begin{enumerate}
  \item Users begin by entering a prompt. Prompts are free-form and unconstrained, reflecting real-world usage rather than predefined tasks. To reduce the blank-page effect, the interface also provides optional prompt suggestions (for now only in the French version of the platform) that users can reuse or adapt, although they are used for less than 6\% of queries (platform analytics; snapshot \snapshotdate).

  \item Two models then generate responses to the prompt. The responses are displayed side by side without any identifying information. Users can read both answers and continue the conversation if they wish, which allows the collection of multi-turn interactions.

  \item Feedback can be provided at two levels. Users may react to individual messages, capturing fine-grained judgments, or vote on the overall conversation by selecting the response they prefer. These two mechanisms produce distinct but complementary signals.

  \item After feedback is submitted, the platform reveals the identities of the models and displays associated metadata. This includes model descriptions and additional contextual information intended to support user understanding. Users are also shown an estimation of the environmental impact of the inference, based on the quantity of tokens generated, the architecture and size of the model. For proprietary models, the architecture and size are estimated based on publicly available information. The energy consumption is estimated thanks to a library called Ecologits \parencite{rince2025ecologits}.
\end{enumerate}

\begin{figure}[H]
  \centering
  \begin{subfigure}[t]{0.48\linewidth}
    \centering
    \includegraphics[width=\linewidth]{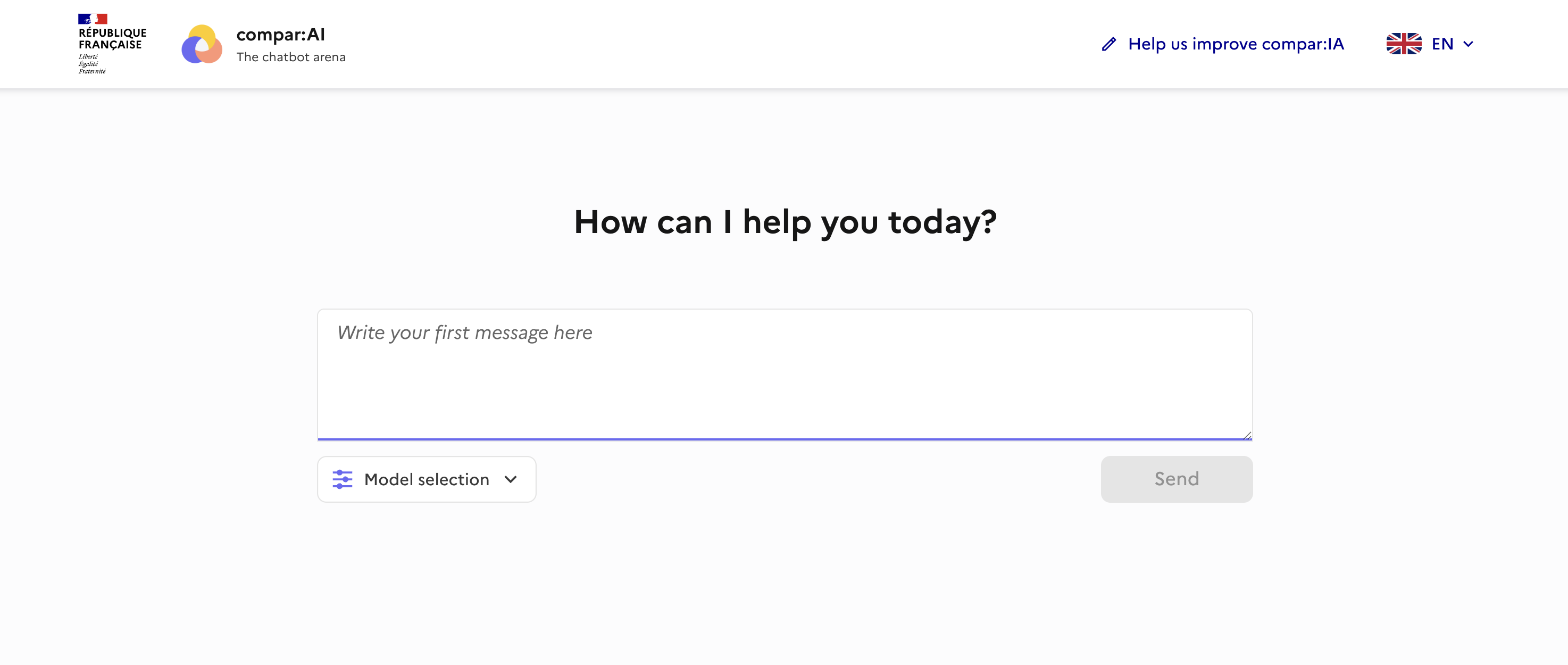}
    \caption{Enter a prompt.}
  \end{subfigure}
  \hfill
  \begin{subfigure}[t]{0.48\linewidth}
    \centering
    \includegraphics[width=\linewidth]{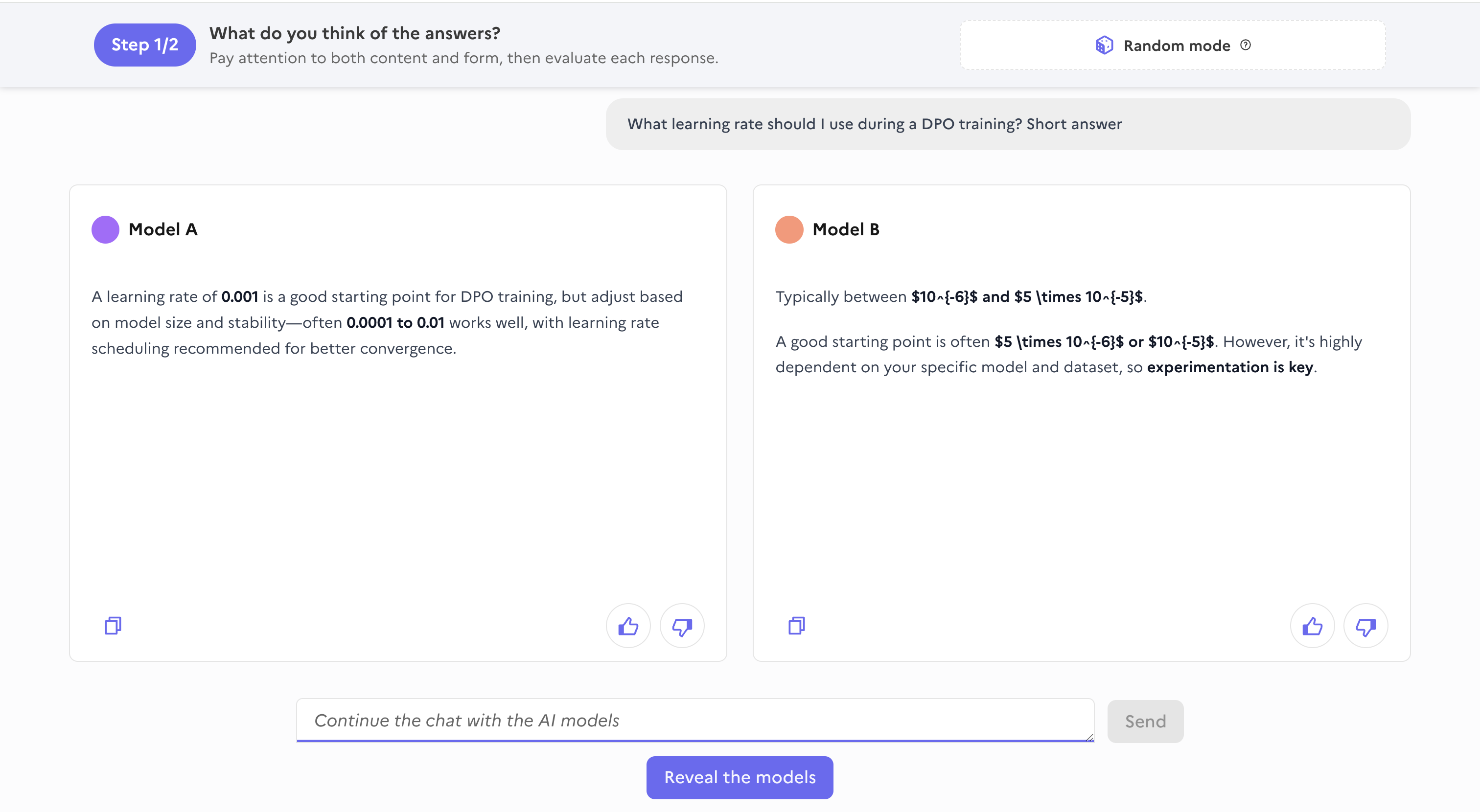}
    \caption{Blind side-by-side responses.}
  \end{subfigure}

  \vspace{0.5em}

  \begin{subfigure}[t]{0.32\linewidth}
    \centering
    \includegraphics[width=\linewidth]{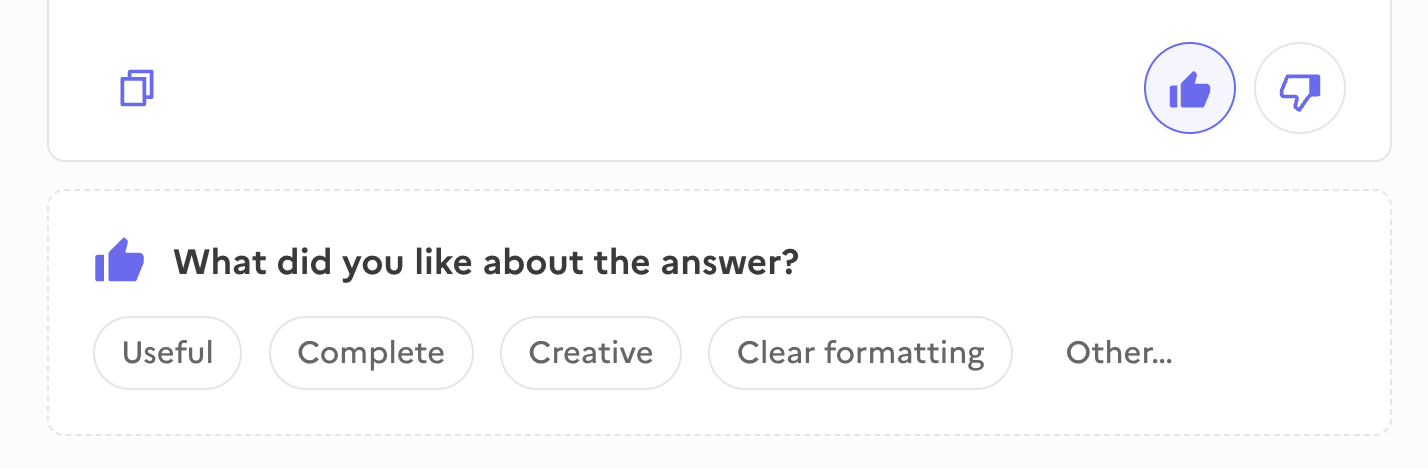}
    \caption{Message reactions.}
  \end{subfigure}
  \hfill
  \begin{subfigure}[t]{0.32\linewidth}
    \centering
    \includegraphics[width=\linewidth]{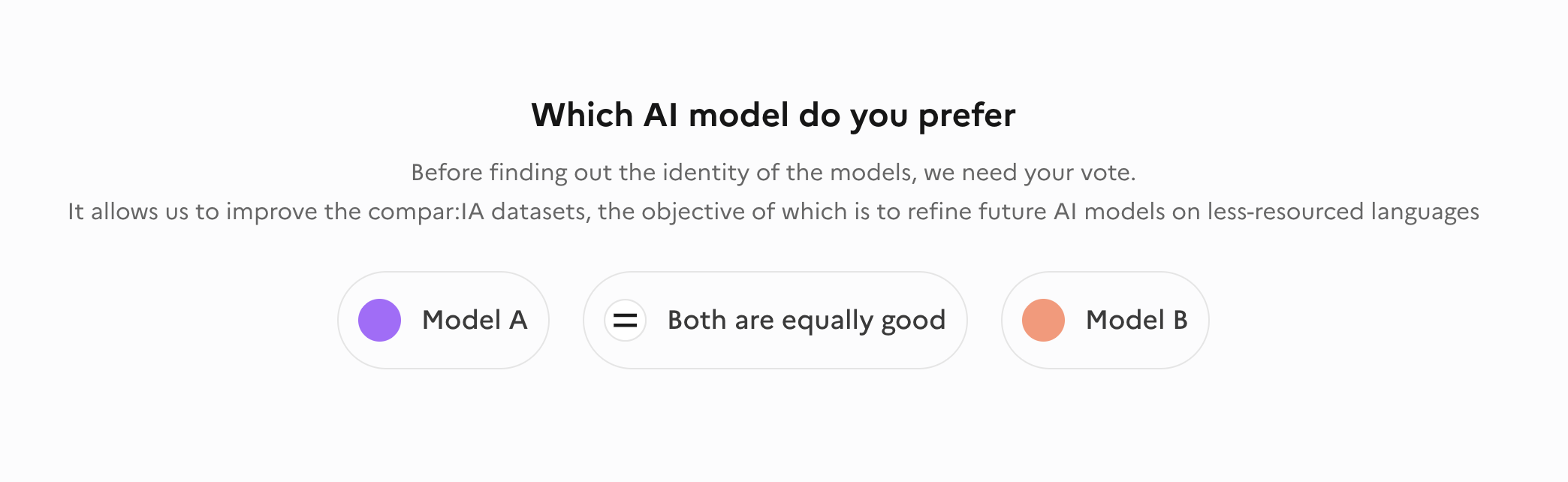}
    \caption{Conversation vote.}
  \end{subfigure}
  \hfill
  \begin{subfigure}[t]{0.32\linewidth}
    \centering
    \includegraphics[width=\linewidth]{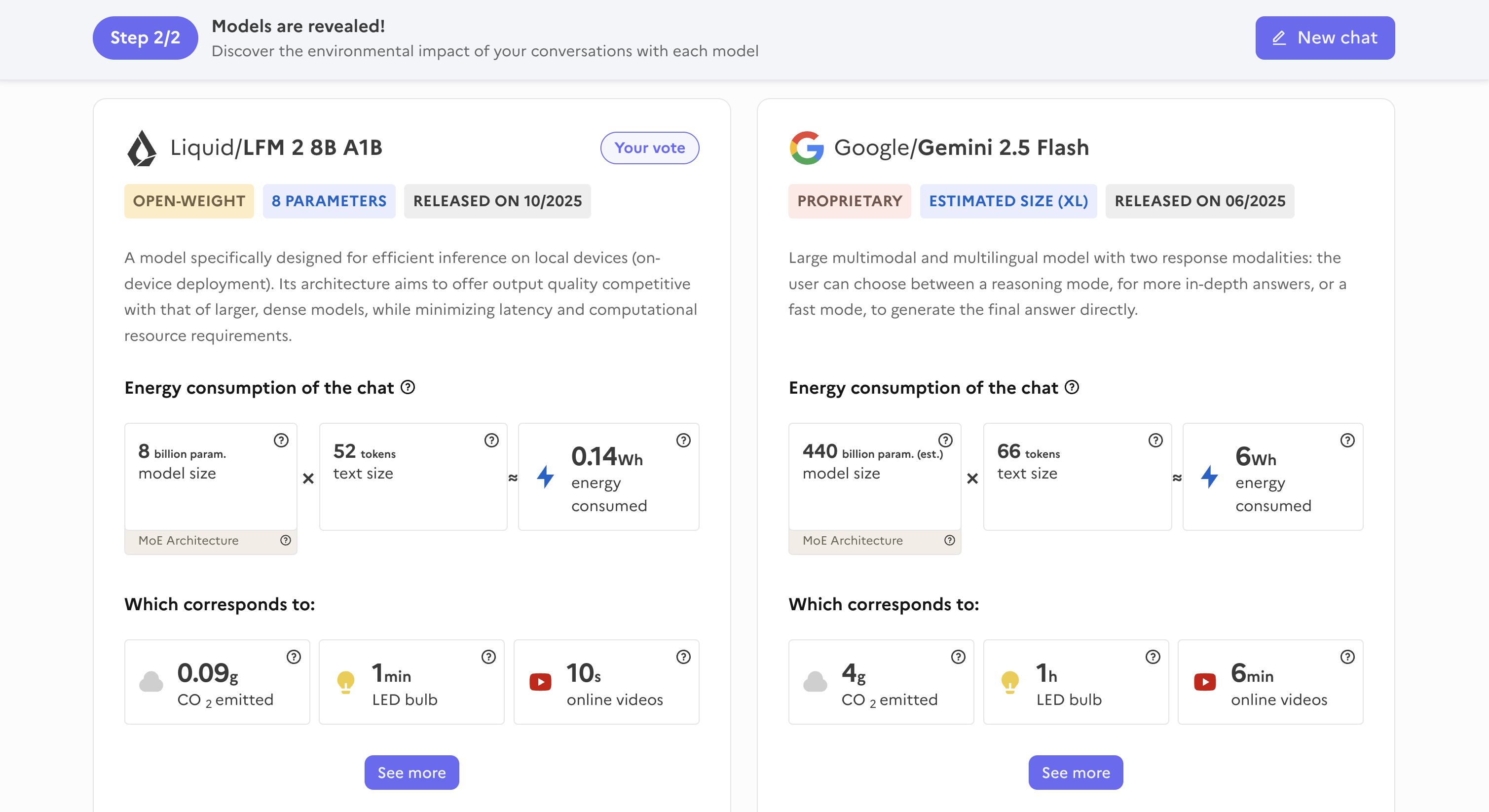}
    \caption{Model reveal and metadata.}
  \end{subfigure}

  \caption{compar:IA user interaction flow.}
\end{figure}

\subsection{Core design principles}

compar:IA maximizes participation by minimizing friction. No account creation is required; users can interact immediately after checking a data reuse consent checkbox. This limits metadata collection but lowers the barrier to entry, especially for non-expert audiences. The platform has been widely used in schools and universities. Optional account creation may be introduced in future iterations, but the baseline experience will remain usable without authentication.

Accessibility is also central. The interface avoids technical terminology, provides contextual explanations, and uses visual cues to guide users. The goal is meaningful participation from users without prior knowledge of LLM architectures or evaluation protocols.

\subsection{Value for end users}

compar:IA gives users free access to diverse language models, including systems less visible in mainstream consumer products. As of \snapshotdate, 104 models (29 proprietary, the rest open-weight or open-source) are available. This diversity challenges the perception that only a few dominant models define the state of the art.

By comparing responses side by side, users observe linguistic, stylistic, and cultural differences between models. Pairwise comparison makes these differences more salient than isolated interactions, highlighting that model behavior is neither uniform nor neutral.

The platform also introduces users to the environmental dimension of AI usage. By associating responses with energy estimates, compar:IA encourages reflection on responsible use without requiring technical expertise.

\subsection{Backend architecture and infrastructure}

The initial version of compar:IA derived from the LMSYS Chatbot Arena codebase and relied on Gradio. Early deployments used inference credits from multiple partners (Scaleway, OVH, Google, Microsoft, Hugging Face). This setup enabled rapid prototyping but was not designed for long-term stability.

As the platform matured, the backend was refactored for sustained public use and higher traffic. The current system uses a FastAPI backend with a SvelteKit frontend, enabling flexible development, improved performance, and finer-grained control over data flows.

In parallel, compar:IA transitioned to a self-financed inference model. Model inference now uses OpenRouter, Hugging Face Inference Providers, or other per-token APIs, allowing integration of a broad model set while maintaining predictable costs and independence from ad hoc credit allocations.

\section{Data collection and dataset construction}

\subsection{Overview of collected data}

compar:IA was opened to the public in October 2024 and has been continuously collecting conversation data since that date.

As of \snapshotdate, the platform has collected over 600,000 free-form prompts, along with more than 250,000 conversation-level preference votes and message-level reactions (snapshot \snapshotdate).

The language distribution is predominantly French (89.14\%). While the platform is technically multilingual and occasionally receives prompts in other languages, French accounts for the large majority of collected data.

\textbf{Top languages:}

\begin{table}[ht]
\centering
\begin{tabular}{lrr}
\toprule
Language & Count & Percentage \\
\midrule
French (fr)  & 357,842 & 89.14\% \\
English (en) & 34,335  & 8.55\% \\
Spanish (es) & 2,571   & 0.64\% \\
Danish (da)  & 1,240   & 0.31\% \\
German (de)  & 1,067   & 0.27\% \\
Italian (it) & 728     & 0.18\% \\
Chinese (zh) & 563     & 0.14\% \\
Arabic (ar)  & 384     & 0.10\% \\
Latin (la)   & 259     & 0.06\% \\
Portuguese (pt) & 234  & 0.06\% \\
\bottomrule
\end{tabular}
\caption{Top languages in compar:IA prompts (as of \snapshotdate).}
\end{table}

Topic wise, technical/educational prompts dominate the distribution (Natural Science + Education = 31.95\%), but other topics are covered as well. This might be due to an overrepresentation of compar:IA's usage in the educational sector.

\textbf{Top categories (15 most frequent):}

\begin{table}[ht]
\centering
\begin{tabular}{lrr}
\toprule
Category & Count & Percentage \\
\midrule
Natural Science \& Formal Science \& Technology & 145,357 & 17.67\% \\
Education & 117,452 & 14.28\% \\
Business \& Economics \& Finance & 83,111 & 10.10\% \\
Society \& Social Issues \& Human Rights & 57,065 & 6.94\% \\
Entertainment \& Travel \& Hobby & 55,244 & 6.72\% \\
Politics \& Government & 52,915 & 6.43\% \\
Culture \& Cultural geography & 44,712 & 5.43\% \\
Arts & 38,937 & 4.73\% \\
Personal Development \& Human Resources \& Career & 36,889 & 4.48\% \\
Other & 30,005 & 3.65\% \\
Law \& Justice & 28,624 & 3.48\% \\
Health \& Wellness \& Medicine & 26,303 & 3.20\% \\
Environment & 26,093 & 3.17\% \\
Daily Life \& Home \& Lifestyle & 23,307 & 2.83\% \\
Food \& Drink \& Cooking & 20,658 & 2.51\% \\
\bottomrule
\end{tabular}
\caption{Top prompt categories (as of \snapshotdate).}
\end{table}

A startup called Bunka.ai has reused a subset of the dataset in order to do some mapping of all the different topics discussed on the platform. Using unsupervised topic modeling and large-scale LLM-based classification, the study examined multiple interaction dimensions, including topics, tasks, language complexity, emotional engagement, and human--AI interaction modes. The results show that French users primarily engage with conversational AI for learning, advice-seeking, content creation, and information retrieval, and that interactions are predominantly augmentative rather than fully automative. A blog post describing the study and methods is available online.\footnote{Hugging Face blog post: ``French people using AI'' (compar:IA $\times$ Bunka.ai study). \url{https://huggingface.co/blog/comparIA/french-people-using-ai}}

\begin{figure}[t]
  \centering
  \includegraphics[width=0.95\linewidth]{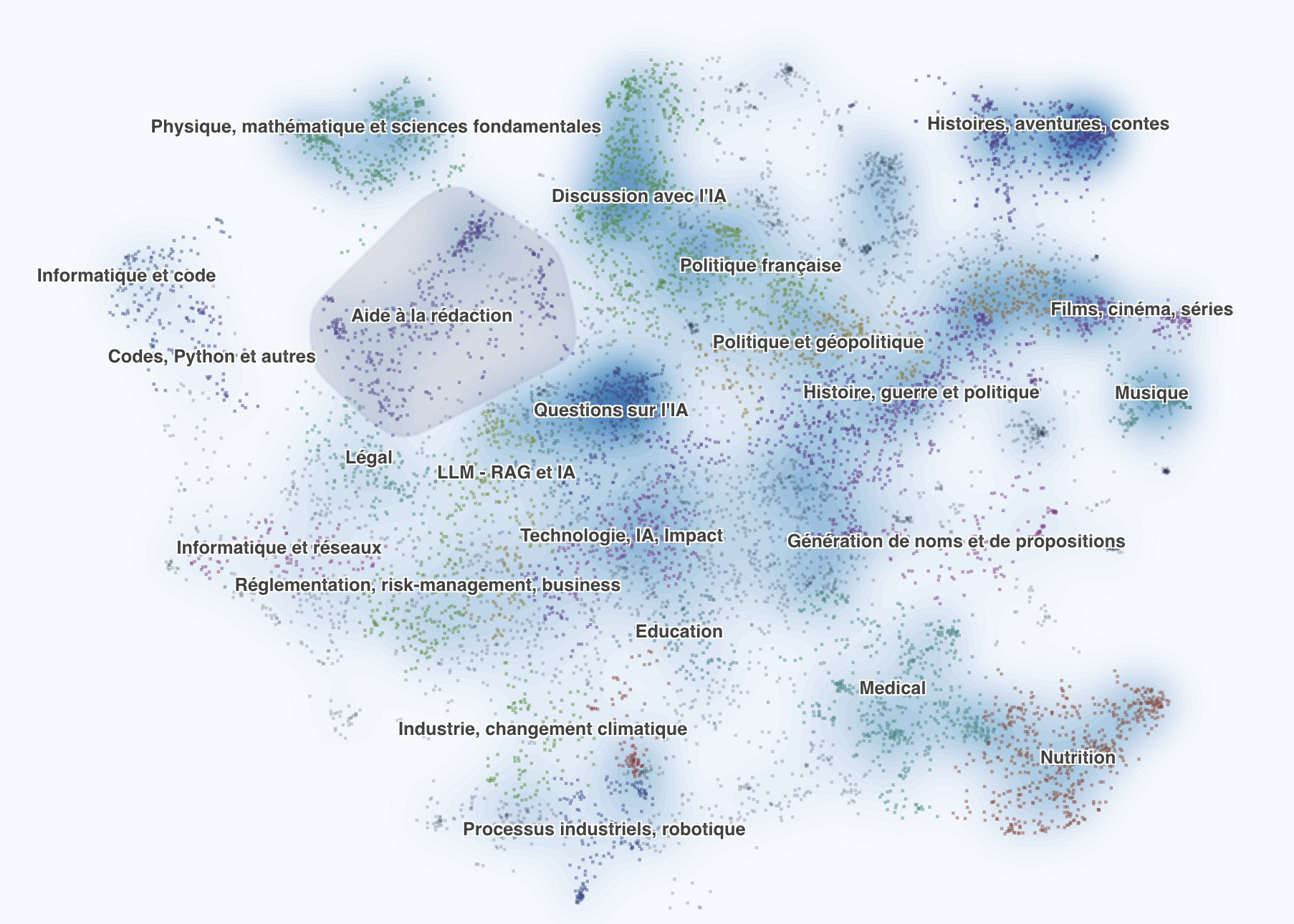}
  \caption{Thematic map of conversational AI uses on compar:IA.}
\end{figure}

\subsection{Published datasets}

Data collected through compar:IA is published as three distinct datasets, each corresponding to a different level of interaction:

\begin{itemize}
  \item \textbf{comparia-conversations}, containing prompts and model-generated responses organized as multi-turn conversations\footnote{Hugging Face dataset: ministere-culture/comparia-conversations. \url{https://huggingface.co/datasets/ministere-culture/comparia-conversations}}
  \item \textbf{comparia-votes}, containing conversation-level pairwise preference annotations\footnote{Hugging Face dataset: ministere-culture/comparia-votes. \url{https://huggingface.co/datasets/ministere-culture/comparia-votes}}
  \item \textbf{comparia-reactions}, containing message-level feedback signals\footnote{Hugging Face dataset: ministere-culture/comparia-reactions. \url{https://huggingface.co/datasets/ministere-culture/comparia-reactions}}
\end{itemize}

The datasets are hosted on Hugging Face and mirrored on data.gouv.fr (\url{https://www.data.gouv.fr/datasets/compar-ia}) to ensure long-term accessibility.

All datasets are released under the Etalab 2.0 open license, standard for French government data. However, responses from proprietary models and some open-weight models are restricted to analysis and evaluation, not training or fine-tuning.

\subsection{Privacy and data filtering pipeline}

compar:IA does not require accounts and does not collect explicit personal data such as names, email addresses, or identifiers (other than short-term session IDs). This simplifies participation but shifts data protection to post-collection filtering.

Before publication, all data pass through an LLM-based personal data detection pipeline. Each conversation is evaluated to determine whether it likely contains personal or sensitive information.

Should a conversation be wrongly identified as not containing PII or other sensitive data, a publicly available form\footnote{\url{https://adtk8x51mbw.eu.typeform.com/to/B49aloXZ}} allows reporting these incidents so that the data points can be removed manually by compar:IA maintainers.

The filtering strategy is deliberately conservative. When personal data is detected, the entire conversation and its associated votes and reactions are excluded, resulting in about 5\% of conversations being filtered out. No attempt is made to mask or anonymize specific spans of text. The raw dataset is uploaded to Hugging Face but gated for research purposes only.

This reduces data volume but limits re-identification risk and simplifies compliance with GDPR and national regulation. The trade-off favors privacy over maximal dataset size. Alternative approaches involving span-level anonymization are under investigation, but the associated risks---incomplete masking, semantic distortion, or residual identifiability---remain significant. Until these can be reliably mitigated, full-conversation exclusion remains the default policy.

\subsection{Comparison with existing datasets}

Methodologically, compar:IA resembles LMarena preference datasets, notably LMSYS Chat-1M\footnote{\url{https://huggingface.co/datasets/lmsys/lmsys-chat-1m}} and newer Arena datasets.\footnote{\url{https://huggingface.co/lmarena-ai/datasets}} However, the linguistic composition differs substantially.

While French represents only a small fraction of conversations in existing open preference datasets (reported at 1.5\% in LMSYS Chat-1M \parencite{chiang2024arena}), compar:IA provides several hundred thousand French-language prompts and interactions.

Other corpora in less resourced languages will be published shortly, as the service expands to these countries, starting with Denmark.

The compar:IA datasets are also enriched with electricity consumption estimates (kwh), based on the Ecologits calculation method.

\section{Adoption, usage, and impact}

\subsection{Platform usage metrics}

Since its public launch in October 2024, compar:IA has attracted a large and steady audience. As of \snapshotdate, the platform has recorded more than 300,000 unique visitors (platform analytics; snapshot \snapshotdate). Usage is not restricted to specific campaigns or events and reflects continuous organic traffic.

Participation is entirely voluntary. As of January 2026 users do not receive compensation, rankings, or badges, and are not required to create accounts. The absence of explicit incentives reinforces the interpretation of collected data as reflective of spontaneous public engagement rather than task-driven annotation behavior.

\subsection{Dataset impact indicators}

Given the difficulty of observing downstream reuse, dataset size serves as a first-order proxy for impact. The number of prompts, conversations, and preference votes indicates both platform adoption and potential utility of the published datasets.

Access metrics on hosting platforms provide partial visibility into reuse. 778 unique users requested access across the three datasets on Hugging Face. On data.gouv.fr the datasets are not gated.

To gather more data about dataset reuse, the compar:IA team sent a survey in October 2025 to people that requested access to the dataset. 25 people responded. They self declared as 68\% of academics/researchers and 20\% representing private companies. They used compar:IA datasets mostly to do model training (32\%), research (24\%), or evaluation (20\%). For example for studying human preferences, language use, prompts and discourse.

This suggests active interest, but systematic reporting of reuse remains limited, particularly among industrial actors, making impact measurement challenging.

\subsection{Mediation, outreach, and growth strategies}

To sustain participation and broaden its audience, compar:IA has been integrated into a range of educational formats. These include workshops, public talks, and educational programs focused on digital literacy and generative AI. In these contexts, the platform is used both as a demonstration tool and as an interactive exercise. For example, PIX, a national online platform open to everyone to assess, develop, and certify digital skills, has integrated compar:IA into its AI curriculum. In 2026, more than 1.5 million students are expected to use compar:IA through this program.

A dedicated mediation format, \emph{Les Duels de l'IA}, was developed to support using compar:IA for education. This format structures interactions around collective discussion about the model answers and specifications, encouraging critical reflection. More than 1400 potential facilitators have submitted their registration to receive material for the workshop. Several hundred people have already submitted the post-event feedback forms.

\subsection{First-party dataset reuses}

Beyond data collection, compar:IA produces secondary outputs derived from preference data.

\subsubsection{Model leaderboard}

compar:IA released a first model leaderboard based on aggregated pairwise preferences.\footnote{\url{https://comparia.beta.gouv.fr/ranking}}\footnote{Hugging Face blog post: ``Publication du premier classement'' (compar:IA leaderboard). \url{https://huggingface.co/blog/comparIA/publication-du-premier-classement}}

The leaderboard was developed in collaboration with PEReN and was made public in November 2025. Rankings are updated on a weekly basis to account for newly collected data.

\begin{figure}[t]
  \centering
  \includegraphics[width=0.95\linewidth]{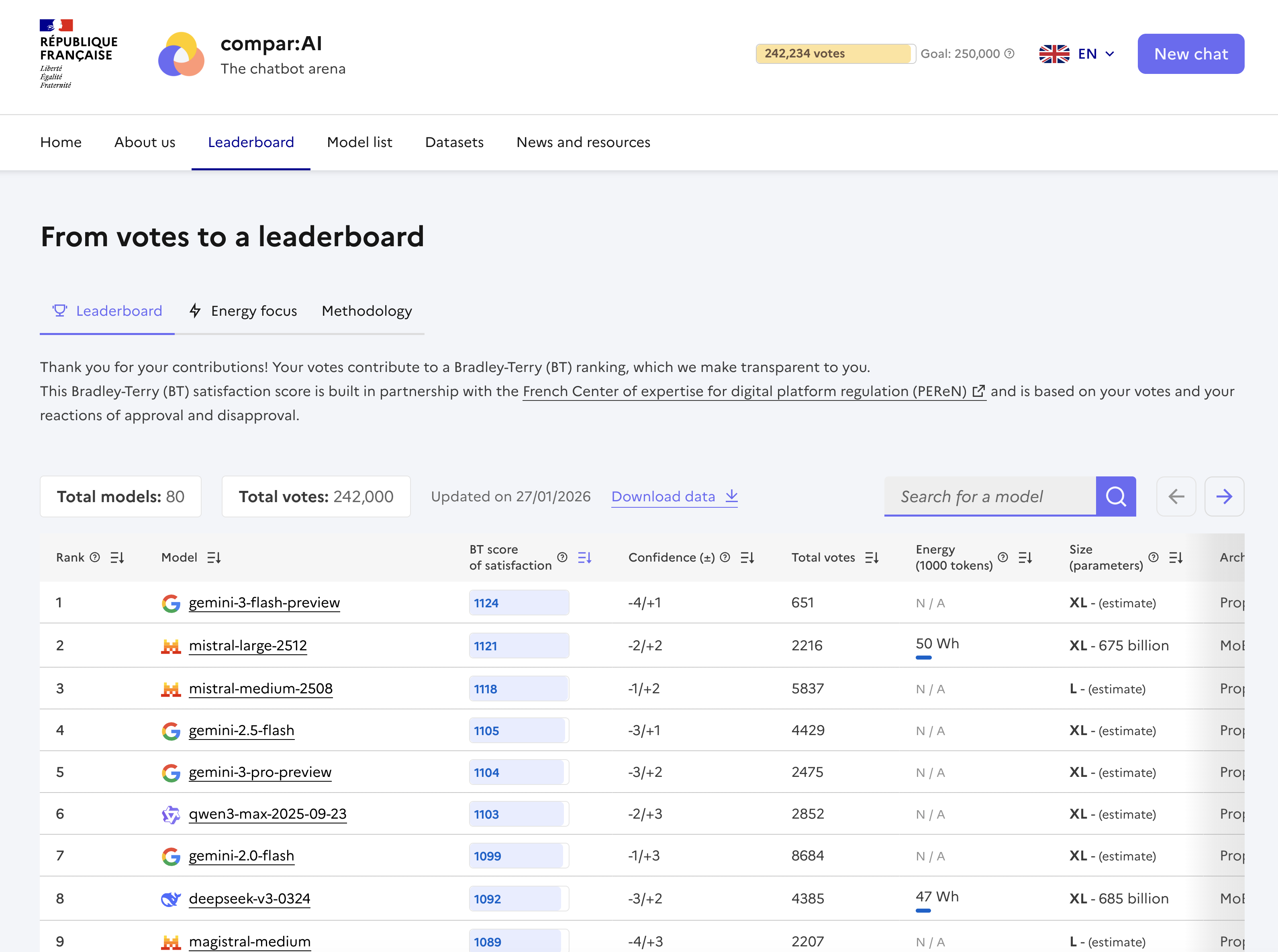}
  \caption{compar:IA model leaderboard (as of \snapshotdate).}
  \label{fig:leaderboard}
\end{figure}

The ranking methodology relies on preference modeling techniques commonly used in pairwise comparison settings, specifically Bradley--Terry models \parencite{bradley1952rank}. The leaderboard is constructed from conversation-level votes and message-level reactions and is designed to reflect relative user preferences rather than task-specific performance.

This output has known limitations. Preference data is influenced by prompt distribution, user population, and self-selection effects. Leaderboard positions should be interpreted as indicative rather than definitive. The primary role is exploratory and educational, not a formal benchmark.

\subsubsection{Thematic analysis of prompt usage}

A collaborative analysis with Bunka.ai was conducted on a subset of over 175,000 compar:IA conversations to characterize prompt usage patterns. Using unsupervised topic modeling and classification, the study identified four dominant interaction types: learning, advice seeking, content generation, and information retrieval, across domains including technology, education, work, health, and culture.

The analysis revealed systematic associations between domains and interaction types. Health-related prompts are predominantly advice-oriented, scientific topics are mainly learning-focused, and creative domains emphasize content generation. Across most domains, interactions are primarily augmentative rather than fully automative, suggesting that users treat conversational AI as an assistive system rather than a replacement for human effort.

The study highlights the value of large-scale, unconstrained conversational data for usage analysis and its limitations, including the absence of multimodal signals and potential prompt selection bias. The full analysis is available online.\footnote{Hugging Face blog post: ``French people using AI''. \url{https://huggingface.co/blog/comparIA/french-people-using-ai}}

\section{Use cases for the AI ecosystem}

\subsection{Research and model development}

The datasets are primarily intended for research and model development workflows relying on human prompt and preference data.

One direct application is preference-based training, including reinforcement learning from human feedback and direct preference optimization \parencite{christiano2017preferences,ouyang2022instructgpt,rafailov2023dpo}.

The prompt and conversation data can also be used as a basis for synthetic data generation. For example, prompts may serve as seeds for controlled generation of additional training data.

\subsection{Studying model usage}

Prompt collections can be analyzed to study real-world usage distributions, topic prevalence, and interaction styles in French conversational AI, supporting sociological research.

compar:IA data can contribute to multilingual evaluation benchmarks. By sampling prompts and preferences, researchers can construct test sets grounded in actual user behavior rather than expert-designed tasks, particularly relevant for underrepresented languages.

\section{Governance and development history}

\subsection{Project genesis and development}

compar:IA originated as an intrapreneurial initiative (commonly referred to as a ``State startup'') within the French public administration, jointly supported by the Ministry of Culture and the Interministerial Directorate for Digital Affairs (DINUM), with the initial problem to be solved being ``how to facilitate access to data in French for training language models while respecting copyright''.\footnote{\url{https://beta.gouv.fr/approche/}}

Early development followed an agile, lean startup approach. Rather than defining a fixed product upfront, the project began with workshops involving public servants, researchers, practitioners, and civil society actors to identify user needs, test prototypes, and surface usability constraints.

The platform evolved iteratively. Initial prototypes validated the core interaction model; subsequent iterations improved interface design, model integration, and data pipelines. Over time, compar:IA transitioned from an experimental prototype to a stable national digital public service.

\subsection{Institutional positioning and objectives}

compar:IA is jointly operated by the Ministry of Culture and DINUM as a non-commercial digital public service.

The platform pursues two complementary objectives: raising public awareness of LLMs (their diversity, biases, and environmental impacts), and collecting human prompt and preference data in French for open release to academic, industrial, and public-sector actors.

Since November 2025, compar:IA is recognised as a digital public good by the Digital Public Goods Alliance.\footnote{DPGA registry entry: \url{https://www.digitalpublicgoods.net/r/comparia}} The platform is free, open-source, and publishes datasets under open licenses. It is non-commercial and does not monetize user activity; success is evaluated through participation, awareness, dataset quality, and downstream reuse. This approach aligns with recent calls to build an open ecosystem for human feedback on AI systems, drawing from peer-production, open-source, and citizen-science practices \parencite{donyehiya2025openfeedback}.

As of writing, in January 2026, compar:IA is in the process of expanding the platform to other European languages. At the moment, the effort is based on bilateral partnerships, but in the medium-term the aim is to develop a digital common with a shared governance.

\section{Limitations}

\subsection{User representativeness}

compar:IA does not collect socio-demographic information. This follows from avoiding account creation and minimizing data collection to lower participation barriers and reduce privacy risks.

As a result, the user population cannot be characterized by age, profession, education level, or geographic distribution beyond coarse language-level inference. This limits preference analysis across user groups and prevents weighting schemes that would correct for population imbalances.

The absence of such metadata constrains bias analysis. While model-level differences can be observed, whether these are driven by specific user subpopulations or usage contexts cannot be determined.

\subsection{Professional and task-specific coverage}

The dataset likely underrepresents professional and sensitive use cases. Because all prompts may be published openly, users are often reluctant to submit work-related, confidential, or regulated queries.

This affects both prompt diversity and applicability to professional tasks. Domains such as law, healthcare, internal administration, or corporate decision-making are sparsely represented compared to general informational or creative prompts.

This also impacts the model leaderboard. Preference rankings from general-public usage may not reflect professional performance and should not be interpreted as task-specific evaluations.

\subsection{Evaluation bias and self-selection}

Participation is voluntary and self-selected. The user population skews toward individuals with existing interest in AI or digital tools. Outreach through educational and institutional networks reinforces this profile.

This introduces evaluation bias. Preferences may not generalize to the broader population of conversational AI users, particularly those with lower digital literacy or different usage patterns.

\subsection{Arena-style platform limitations}

Arena-style evaluation platforms constitute a distinct usage context that differs from everyday chatbot interaction. Users on such platforms may adopt an evaluative or experimental mindset, submitting shorter, more simplified prompts designed to ``test'' models rather than to accomplish real tasks. This can lead to shorter conversations and different prompt distributions compared to naturalistic usage.

Additionally, the communication and framing around compar:IA may overly encourage culture-specific or French-language-oriented questions, which could overrepresent certain prompt types relative to how French speakers actually use conversational AI in daily life.

More fundamentally, pairwise evaluation itself has structural limits. Comparing two responses highlights relative differences but may obscure absolute quality or fail to capture dimensions that are not easily contrasted, such as factual completeness, long-term usefulness, or safety considerations. Certain model behaviors may therefore be under-emphasized or ignored by the evaluation setup, depending on the prompt and the comparison context.

Recent work has begun examining methodological aspects of arena-style evaluation more broadly \parencite{singh2025leaderboardillusion}. While compar:IA contributes to transparency by releasing all collected data openly, it shares some inherent characteristics of pairwise evaluation platforms that warrant continued methodological attention.

\subsection{Model configuration and infrastructure constraints}

Several technical factors affect the comparability of models on the platform.

First, \textbf{system prompt policies} have evolved over time. Open-weight models served through the platform initially operated without system prompts, while proprietary models typically include default system instructions. This asymmetry may have disadvantaged open models in certain contexts. The issue became particularly apparent when integrating specific models (e.g., the ``Chocolatine'' model), which required a custom system prompt to perform comparably. The platform has since adopted more consistent system prompt policies, but historical data may reflect these inconsistencies.

Second, \textbf{closed-source model behavior is not fully transparent}. When querying proprietary models through APIs, we cannot verify whether the response comes solely from the language model or involves additional preprocessing, post-processing, routing logic, or tool use. We have attempted to avoid serving models known to perform tool calling or internet access, but complete certainty is not possible for closed-source systems.

Third, \textbf{model quantization varies across inference providers}. While the platform targets full-precision or high-quality quantized model variants, some inference providers (notably through OpenRouter) may serve lower-precision quantized versions (e.g., 8-bit or lower) without explicit disclosure. We have attempted to avoid heavily quantized variants, but provider-level decisions are not always controllable or visible to the platform operators.

These infrastructure-level factors introduce potential confounds in model comparisons and should be considered when interpreting leaderboard rankings or preference distributions.

\subsection{Other potential biases}

Beyond the limitations discussed above, several additional sources of bias may affect the collected preference data.

First, \textbf{answer style} can bias preferences. Models that produce longer, more confident, with more emojis, or more conversational responses may be favored over more concise or cautious ones, even when the underlying information quality is similar. This stylistic bias is inherent to open-ended, unconstrained prompts and is difficult to disentangle from genuine user preference.

While preference data is essential, it is also known to be prone to subtle and systematic biases. Common effects include \emph{prefix bias}, where early parts of a completion disproportionately influence preference judgments \parencite{kumar2025prefixbias}, which can then propagate to downstream models \parencite{bharadwaj2025flattery}. Other well-documented phenomena include \emph{sycophancy} \parencite{sharma2024sycophancy}, as well as \emph{verbosity and length-related biases} \parencite{singhal2023length,bu2025adaptive} and \emph{formatting biases} \parencite{zhang2025formatbias}.

Many of these effects reflect underlying human judgment tendencies rather than annotation errors and are therefore difficult to eliminate at collection time. By releasing large-scale preference data openly, compar:IA aims to enable the research community to study these biases post hoc and to develop causal analyses, filtering strategies, or post-processing methods that help isolate and mitigate such effects in downstream training and evaluation.

Second, \textbf{model latency} can influence user judgments. Faster responses may be perceived as more fluent or reliable, while slower responses may negatively impact user preference independently of content quality. Although responses are displayed simultaneously once generated, differences in generation time may still affect user attention and engagement.

Third, \textbf{user voting behavior is not fully observable}. Some users may not read both responses carefully before voting, or may rely on superficial cues such as response length, tone, or formatting. While this behavior reflects real-world usage patterns, it introduces noise into the preference signal and reduces the reliability of individual votes.

These biases do not invalidate the collected data but should be considered when interpreting preference signals and downstream analyses.

\section{Future directions}

\subsection{Internationalisation and multilingual expansion}

A primary direction is extending compar:IA to additional languages. While currently French-dominant, the underlying architecture is language-agnostic and could be reused for any language.

Current expansion efforts focus on European languages that are underrepresented in existing evaluation datasets. For each new language, the objective is to reach dataset sizes sufficient for meaningful reuse, with a target on the order of tens of thousands of prompts and preference votes per language. Beyond Europe, the open-source infrastructure could support deployment in other linguistic contexts where preference data remains scarce.

Multilingual deployment also enables cross-linguistic analysis, studying how model behavior and user preferences vary by linguistic and cultural context using consistent protocols.

\subsection{Enhanced data annotation and metadata}

Future iterations may introduce optional metadata collection, including broad user categories or self-declared familiarity with AI systems.

Such metadata would be opt-in and privacy-preserving, enabling contextualized preference signals and basic stratified analyses rather than fine-grained profiling.

\subsection{Specialized arenas}

Another direction involves specialized arenas for professional or sectoral use cases. Unlike the general-public platform, these could operate with controlled cohorts, explicit consent, and stricter access conditions.

Specialized arenas would enable higher-quality, task-specific preference data in domains like public administration, education, or regulated professions, allowing more reliable interpretation of preference rankings within defined contexts.

\section{Conclusion}

This paper presented compar:IA, a public LLM arena designed to collect large-scale human preference data in French through blind pairwise comparison. We described the platform's institutional context, design principles, and user interaction model, as well as the structure, publication, and early impact of the resulting datasets. With more than 600,000 prompts and over 250,000 preference votes released under open licenses, compar:IA constitutes a significant contribution to the currently limited pool of open, non-English preference data for conversational AI.

Beyond data volume, compar:IA demonstrates that large-scale human evaluation can be carried out with a general public when interfaces are accessible and participation friction is minimized. The platform shows that preference-based evaluation is not limited to expert annotators or proprietary systems, and that real-world prompts can be collected responsibly at scale within a public-sector context.

The project highlights the role public institutions can play in AI evaluation infrastructure. By operating outside commercial incentives and prioritizing openness, compar:IA supports collective learning across academic, industrial, and public-sector actors. Its governance illustrates how public administrations can balance participation, transparency, and privacy in AI-related services.

Finally, compar:IA offers a replicable model for language-specific, human-centered evaluation. Its focus on linguistic diversity and cultural context complements existing English-centric benchmarks. As multilingual extensions develop, compar:IA may serve as a foundation for open evaluation infrastructures that better reflect the linguistic and cultural plurality of AI users.

\section*{Acknowledgments}

This work benefited from the support, advice, and resources of many individuals and organisations.

\textbf{Institutional and operational support.}
We gratefully acknowledge the French Ministry of Culture, in particular Mathilde Bras, Ned Baldessin, Léo Wellhoff, Romain Delassus, and Guillaume Combe, as well as DINUM, notably Pierre Pezziardi and Elsa Le Duigou, for sustained support and guidance throughout the project.

\textbf{Scientific, technical, and strategic advice.}
We thank colleagues, partners, and organisations for feedback, expertise, discussions, and exchanges throughout the project, including Inria (Djamé Seddah, Benoît Sagot), CNRS and IDRIS (Patrick Paroubek, Maziyar Panahi), GENCI and the Jean Zay infrastructure, the Bibliothèque nationale de France (BnF), INA, Linagora (Michel Marie Maudet, Julie Hunter), ALEIA (Antoine Couret), Mistral AI (Guillaume Lample), Illuin (Manuel Faysse, Gautier Viaud), ReciTAL (Gilles Moyse), OPSCI (Emile Hazard, Clément Bénesse), Le Voice Lab (Karel Bourgois), Teklia (Christopher Kermorvant), PLEIAS (Anastasia Stasenko, Pierre-Carl Langlais), ELDA / ELRA (Khalid Choukri), the Cité européenne des scénaristes (Pauline Rocafull), Dawex, Ouest-France, Google (Robert Dadashi, Léonard Hussenot), La Javaness, the Pantagruel project, Hugging Face (Clément Delangue, Clémentine Fourrier, Daniel van Strien, Nathan Habib, Quentin Lhoest), Meta (Thomas Mesnard), Liquid AI (Maxime Labonne), Cohere (Julia Kreutzer, Shivalika Singh), AI21 (Johanna Kramer), Hugues de Mazancourt, make.org, PIX (Marie Bancal, Benjamin Marteau), Datactivist (Samuel Goëta, Loup Cellard, Laurane Coudriet), Latitudes (Margaux Levisalles, Mélanie Brisard, Pauline Mélédo), Galances Conseil (Jérôme Lucereau, Nicolas Blanchon), CLEMI, Réseau Canopé, CAIRE, UNESCO, Mednum (Ondine Vernier), DRANE PACA, DRANE Strasbourg, the Conseil de l'IA et du numérique (Jean Cattan, Joséphine Corcoral, Cécile Ravaux, Magali Jacquemet, Gabriel Ertle), the Campus du Numérique Public (Aude Chouleur, Sophie Louet, Marie Charbonnel), Médialab Sciences Po (Sylvain Parasie, Valentin Goujon), and CREIA (Clément Fantoli).

\textbf{Funding and public support.}
This project was funded by DINUM and the French Ministry of Culture.

\textbf{Data reuse and downstream applications.}
We acknowledge PEReN and Bunka.ai for reusing and building upon the datasets produced as part of this work. We also thank researchers, independent developers, and AI labs who have reused the datasets and shared feedback with us, even when such reuse has not been publicly disclosed.

\textbf{Compute and inference resources.}
We thank Google, Microsoft, Hugging Face, Scaleway, and OVH for providing inference credits and computational resources in the initial stages of the project.

\textbf{Pre-print review.}
We thank David Salinas (Group Lead, OpenEuroLLM) and Kenneth Enevoldsen (Aarhus University, Denmark) for their valuable reviews of this pre-print.

\printbibliography

\end{document}